%% file: tarau_iclp23.tex
\documentclass[submission,copyright,creativecommons]{eptcs}
\providecommand{\event}{ICLP 2022} 
\usepackage{underscore}           

\usepackage{amssymb}
\usepackage{hyperref}
\usepackage{booktabs}
\usepackage[english]{babel}
\input pheader22.tex

\usepackage{breakurl} 

\begin{document}

\title{ 
Natlog:  Embedding Logic Programming into the Python Deep-Learning Ecosystem
}
\author{Paul Tarau
\institute{University of North Texas
}
\email{paul.tarau@unt.edu}
}

\def \authorrunning{Paul Tarau}
\def\titlerunning{Natlog:  Embedding Logic Programming into the Python Deep-Learning Ecosystem}

\date{}
\maketitle

\begin{abstract}
Driven by expressiveness commonalities of Python and our Python-based embedded logic-based language Natlog, we design high-level interaction patterns between equivalent language constructs and data types on the two sides.

By directly connecting generators and backtracking,  nested tuples and terms,  coroutines and first-class logic engines, reflection and meta-interpretation, we enable logic-based language constructs to access the full power of the Python ecosystem. 

We show the effectiveness of our design  via  Natlog apps working as orchestrators for JAX and Pytorch pipelines and  as DCG-driven GPT3 and DALL.E prompt generators.

\vskip 0.2cm
\noindent
{\bf Keyphrases}: {\em
embedding of logic programming in the Python ecosystem,
high-level inter-paradigm data exchanges,
coroutining with logic engines,
logic-based neuro-symbolic computing,
logic grammars as prompt-generators for Large Language Models,
logic-based neural network configuration and training.
}
\end{abstract}


\section{Introduction}

Turing-complete programming languages win users based on how easily one can make computers do things, for which we will use here {\em expressiveness} as an umbrella concept. Declarative programming has been seen over the years as an {\em expressiveness enhancer}, assuming that telling {\em what} rather than {\em  how} to do things makes it easier to achieve a desired outcome. 
With this view of declarative programming in mind, besides competition from today's functional programming languages and proof assistants, 
all with strong declarative claims, logic-based languages face an even stiffer competition from the more radical approach coming from deep learning.

To state it simply, this manifests as replacement of rule-based, symbolic encoding of intelligent behaviors via machine learning, including  unsupervised learning among which  transformers \cite{transfo} trained on large language models have achieved outstanding performance in fields ranging from natural language processing to computational chemistry and image processing.
For instance, results in natural language processing with prompt-driven generative models like GPT3 \cite{gpt3} or prompt-to-image generators like DALL.E \cite{dalle} or Stable Diffusion \cite{stabdif} have  outclassed similarly directed symbolic efforts. In fact, it is quite hard to claim that a conventional programming language (including a logic-based one) is as declarative as entering a short prompt sentence describing a picture and getting it back in a few seconds.

Thus it is becoming clearer as time goes by, that ownership of the declarative umbrella is slowly transitioning to deep neural networks-based machine learning tools that replace human coding (including that done in declarative languages) with models directly extracted from labeled and more and more often from raw, unlabeled data.

This forces us to question, as lucidly as possible, what contribution logic-based reasoning and more concretely logic-based language design  can bring to this fast evolving ecosystem, and what language features are needed to enable them.

\subsubsection*{Language features facilitating interoperation of a logic-based language with the Python deep-learning ecosystem}

Rather than fighting it, we  will explore in this paper several ways to join and enhance this strongly disruptive ecosystem. Toward this end, we will design logic-based language features that will not only facilitate embedding of logic-based reasoning in the Python but also simplify, syntactically and semantically the interaction with the deep learning tools that make the ecosystem itself appealing.

While aware that more accurate (but also more intricate) solutions might come out in the near future as part of  emerging  neuro-symbolic computing research\footnote{
a fast evolving research field, thoroughly overviewed in \cite{neurosym}
} we will apply our findings to two mechanisms of interoperation with deep learning tools, of practical interest today, where the embedding of a logic programming language in the ecosystem  brings a fresh approach.

\subsubsection*{Logic languages as orchestrators for deep learning architectures}

Let's start by observing that the actual code base enabling most of today's deep learning systems is  a convoluted, ungeneralizable mix of unstructured scripting, low level GPU-acceleration and linear-algebra libraries.  

This suggests that there's plenty of room to enhance and clarify the design and implementation of the networks themselves in the declarative frameworks logic-based programming languages provide.
To facilitate interoperation with the ecosystem of logic-based orchestrators for deep-learning tools, we will design a component composition framework implemented in a lightweight  logic language embedded in Python.

\subsubsection*{Logic languages as prompt-generators and Large Language Model interaction refiners}
The emergence of Large Language Models like GPT3 and their refinements like ChatGPT and text-to-image variants like DALL.E exhibits a strong dependency on an essentially declarative component: {\em prompt engineering}.
Prompt engineering can be seen as a goal-driven language generation, a logically specifiable endeavor for which we will show that logic grammars  are a natural match.

To make  our research goals and related language design proposals experimentally testable, we have  embedded in Python  a lightweight Prolog-like language, Natlog.
We refer to \cite{iclp21} for a description, in an early, proof-of-concept version, of  its concrete syntactic and semantic features, while we will
focus here on  advanced use cases based on a fresh ``from-scratch'' reimplementation.

The paper is organized as follows. 
Section \ref{nat} introduces and motivates  the key design ideas behind the Natlog embedded logic-based language, in particular Natlog's coroutining mechanisms using First-Class Logic Engines.
Section \ref{jax} shows uses of Natlog scripts as orchestrators for designing, training and testing deep-learning systems in the JAX and Pytorch ecosystems.
Section \ref{prompts} discusses the use Natlog's logic grammars as prompt generators for text-to-text and text-to-image large language model neural networks.
Section \ref{rel} discusses related work and
section \ref{conc} concludes the paper.

{\em Caveat emptor:} As the paper is about interoperation of language constructs between Python and a Prolog-like logic programming language implemented in it, we assume that the reader is fluent in Prolog and Python. We also assume exposure to essential language constructs of today's very high-level programming languages as well as   familiarity with deep learning frameworks and some key design and implementation decisions behind them.

\section{Key Design Ideas behind Natlog, a Lightweight Prolog-dialect Embedded in Python}\label{nat}
Our Natlog system has been originally introduced in \cite{iclp21}, to which we refer to for syntax, semantics and low level implementation details.
It is currently evolving as a fresh 
implementation\footnote{at \url{https://github.com/ptarau/natlog}, ready to install with ``{\tt pip3 install natlog}''},
and it will be used as a testbed for the key ideas of this paper.

\subsection{Prolog's semantics, but with a lighter syntax}

While keeping Natlog's semantics the same as Prolog's LD-resolution, 
we  have brought its syntax a step closer to  natural language.
In particular, we are not requiring predicate symbols to wrap parenthesized arguments or predicate symbols to be constants.
As a quick glimpse at its syntactic simplifications, here is a short extract from the usual ``family'' program in Natlog syntax.
\begin{code}
sibling of X S: parent of X P, parent of S P, distinct S X.

grand parent of X GP: parent of X P, parent of P GP.
\end{code}

\begin{code}
ancestor of X A : parent of X  P, parent or ancestor P A.

parent or ancestor P P.
parent or ancestor P A : ancestor of P A.
\end{code}

\subsection{A Quick Tour of Natlog's ``Expressiveness Lifters''}

Expressiveness is the relevant distinguishing factor between Turing-complete languages. It can be seen as a pillar of code development automation as clear and compact notation entails that more is delegated to the machine. At the same time, expressiveness enhancers need to be kept as simple as possible, with their users experience in mind.

\subsubsection{A finite function API}

Finite functions (tuples, lists, dictionaries, sets) are instrumental in getting things done with focus on the problem to solve rather than its representation in the language.

In Natlog they are directly borrowed from Python. They can be easily emulated in any Prolog system, but often with a different complexity than if they were natively implemented.

In an immutable form, as well as enabled with backtrackable and non-backtrackable updates, finite functions implemented as dynamic arrays and hash-maps  offer a more flexible and semantically simpler alternative to reliance on Prolog's {\tt assert} and {\tt retract} family of built-ins.

\subsubsection{ Built-ins as functions or generators}

Reversible code like in Prolog's classic {\tt append/3} examples or the use of DCGs in both parsing and generation are nice and unique language features derived from the underlying LD-resolution semantics, but trying to lend reversibility and multi-mode uses to built-ins obscures code and hinders debugging. Keeping Natlog's built-ins uniform and predictable, while not giving up on flexibility, can be achieved by restricting them to:

\BI
\I functions with no meaningful return like {\tt print}, denoted in Natlog by prefixing their Python calls with ``\verb~#~''.
\I functions of  $N$ inputs returning a single output  as the last argument of the corresponding predicate with $N+1$ arguments, denoted in Natlog by prefixing their calls with a backquote symbol ``\verb~`~''. Note that this syntax, more generally, also covers Python's {\em callables} and in particular class objects acting as instance constructors. 

\I generators with  $N$ inputs yielding a series of output values on backtracking by binding the $N+1$-th argument of the corresponding predicate, denoted in Natlog by prefixing their call with two backquotes `` \verb~``~''.

\EI

\subsubsection{Interoperation with Python, as seen from a few Natlog library predicates}

Interaction with Python's finite functions happens via function  and generator calls. For instance, the predicate {\tt argx} implements backtracking over all elements of a tuple or list by relaying on the Python generator {\tt range} that iterates over all index positions from where {\tt arg} picks an item to be unified with variable {\tt X}.
\begin{code}
argx I T X: `len T L, ``range 0 L I, `arg T I X.
\end{code}
Given  Natlog's expected practical uses as a Python package, even when inside Natlog's REPL, answers are shown as the corresponding Python objects,
given the one-to-one correspondence between terms and nested tuples and between variable binding and dictionaries.
Note also that constructors like {\tt dict} (and any other Python callables) are usable directly with the same
syntax as function calls. 
\begin{codex}
?- `dict ((one 1) (two 2) (three 3)) D?
ANSWER: {'D': {'one': 1, 'two': 2, 'three': 3}}

?- eq (a b c) T, argx I T X?
ANSWER: {'T': ('a', 'b', 'c'), 'I': 0, 'X': 'a'}
ANSWER: {'T': ('a', 'b', 'c'), 'I': 1, 'X': 'b'}
ANSWER: {'T': ('a', 'b', 'c'), 'I': 2, 'X': 'c'}
\end{codex}

Natlog uses cons-lists like {\tt (1 (2 (3 ())))} for the usual, unification-based list operations.
A few built-in predicates support their conversion to/from Python tuples or lists:
\begin{code}
to_tuple Xs T : `from_cons_list_as_tuple Xs T.
to_cons_list T Xs : `to_cons_list T Xs.
\end{code}
\subsubsection{Reflecting metaprogramming constructs}
In function and generator calls, Python's {\tt eval} is used to map the Natlog name of a function or generator to its Python definition.
However, to conveniently access object and class attributes, Natlog implements {\tt setprop} and {\tt getprop} relying directly
on corresponding Python built-ins.
\begin{code}
setprop O K V : #setattr O K V.
getprop O K V : `getattr O K V.
\end{code}
Similary, method calls are supported via the Python function {\tt meth\_call} as in the following stack manipulation API:
\begin{code}
stack S : `list S.  
push S X : #meth_call S append (X).
pop S X : `meth_call S pop () X.
\end{code}
In fact, a method call has a surprisingly succinct Python definition, a testimony that {\em elegant metaprogramming constructs on the two sides make language interoperation unusually easy}.
\begin{code}
def meth_call(o, f, xs):
    m = getattr(o, f)
    return m(*xs)
\end{code}
As the reader familiar with Python will notice, a method ``{\tt m}'' is simply an attribute of an object ``{\tt o}'', directly callable once it has been  retrieved from its name ``{\tt f}''.

These predicates are part of the Natlog library code in file {\tt lib.nat}\footnote{at \url{https://github.com/ptarau/natlog/blob/main/natlog/natprogs/lib.nat}} that can be included as part of a Natlog script with help of the Python function {\tt lconsult}\footnote{in file \url{https://github.com/ptarau/natlog/blob/main/natlog/natlog.py} }.

\subsubsection{Reflecting the type system}

As the following examples show, Python's ``{\tt type}'' built-in can be used to reflect and inspect Natlog's data-types.
\begin{codex}
?- `type (a b) T?
ANSWER: {'T': <class 'tuple'>}
?- `type a T1, `type b T2, eq T1 T2.
ANSWER: {'T1': <class 'str'>, 'T2': <class 'str'>}
?- `type X V, eq X a, `type X C?
ANSWER: {'X': 'a', 'V': <class 'natlog.unify.Var'>, 'C': <class 'str'>}
\end{codex}
Note in the last example, that after unification, the type of a variable is dereferenced to the type of its binding.

\subsection{ Natlog's First Class Logic Engines}\label{engs}

Constraint solvers bring to logic-based languages an automated coroutining mechanism when they suspend computations until more data is available and propagate constraints to the inner loops of SLD-derivations with impressive performance gains including on NP-complete problems.
However this implicit coroutining mechanism does not reflect control on backtracking and
does not expose the interpreter itself as a first-class object. 

One can think about First Class Logic Engines as a way to ensure the {\em full  meta-level reflection} of the execution algorithm.
As a result, they enable on-demand computations in an engine rather than the usual eager execution
mechanism of Prolog.

We will spend more time on them as we see them as ``the path not taken''  that
can bring significant expressiveness benefits to  logic-based languages, 
similarly to the way Python's {\tt yield} primitive supports creation of user-defined generators and other compositional asynchronous programming constructs.

\subsubsection{A First-class Logic Engines API}
To obtain the full reflection of Natlog's multiple-answer generation mechanism,
 we will make fresh instances of the interpreter first-class objects.
 
A {\em logic engine} is a Natlog
language processor reflected through an API that 
allows its computations to be controlled 
interactively from another {\em logic engine}.

This is very much the same thing as a programmer controlling Prolog's interactive 
toplevel loop: launch a new goal, ask for a new answer, interpret it, react to it.
The exception is that it is not the programmer, but it is the program that does it!

We will next summarize the  execution mechanism of Natlog's interoperating logic engines.

\noindent The predicate {\tt eng AnswerPattern Goal Engine} works as follows:
\BI
\I creates a new instance of the Natlog
interpreter, uniquely identified by {\tt Engine}
\I shares code
with the currently running program
\I it is initialized
with {\tt Goal} as a starting point, but not started
\I 
{\tt AnswerPattern} ensures that
answers returned by the engine will be instances of the pattern.
\EI
The predicate {\tt ask Engine AnswerInstance} works as follows:
\BI
\I tries to harvest the answer computed from {\tt Goal}, 
as an instance of {\tt AnswerPattern}
\I if an answer
is found, it is returned as {\tt (the AnswerInstance)}, 
otherwise the atom {\tt no} is returned
\I it is used to retrieve successive 
answers generated by an Engine, on demand
\I
it is responsible for actually 
triggering computations in the engine
\EI 

{\em One can see this as transforming
 Natlog's backtracking over all answers
into a deterministic stream of lazily generated answers.
}
\vskip 0.3cm

\noindent Finally, the predicate {\tt stop Engine} works as follows: 
\BI 
\I stops the Engine, reclaiming the resources it has used
\I ensures that {\tt no} is returned for all future queries
\EI
In Natlog these predicates are implemented as built-ins.

\subsubsection{The coroutining Mechanism Implemented by the Engine API}

\paragraph{Natlog's yield operation: a key co-routining primitive}
The annotation ``\verb~^Term~''  extends our coroutining mechanism by allowing
answers to be {\em yield from arbitrary places} in the computation.
It works as follows:
\BI
\I it  saves the state of the engine and transfers 
{\em control} and a {\em result} {\tt Term} to its client
\I the client 
will receive a copy of {\tt Term} simply by using 
its {\tt ask/2} operation
\I an engine returns control to its client 
when initiating a yield operation as 
when a computed answer becomes available.
\EI
As implemented in Python, engines can be seen simply as a special case
of generators that yield one answer at a time, on demand.

We will outline next, with help from a few examples, a few expressiveness improvements First Class Logic Engines can bring to a logic-based programming language.

\subsubsection{An infinite Fibonacci stream with yield}

Like in a non-strict functional language, one can create an infinite recursive loop from which values are yielded as the computation advances:
\begin{code}
fibo N Xs : eng X (slide_fibo 1 1) E,  take N E Xs.

slide_fibo X Y :  with X + Y as Z,  ^X, slide_fibo Y Z.
\end{code}
Note that the infinite loop's results, when seen from the outside, show up as a stream of answers as if produced on backtracking.
With help of the library predicate {\tt take}, we extract the first 5 (seen as a Python dictionary with name ``{\tt X}'' of the variable as a key and the nested tuple representation of Natlog's list as a value), as follows:
\begin{codex}
?- fibo 5 Xs?
ANSWER: {'Xs': (1, (1, (2, (3, (5, ())))))}
\end{codex}

\subsubsection{The {\tt trust} operation}
When the special atom {\tt trust} is yielded, the goal that follows it
replaces the goal of the engine, with all backtracking below that point discarded and
all memory consumed so far made recoverable. As a practical consequence, infinite loops
can work in constant space, even in the absence of last call optimization.

Using it, the predicate {\tt loop} shows how to generate an infinite sequence of natural numbers. 
\begin{code}
loop N N.
loop N X : with N + 1 as M, ^trust loop M X.
\end{code}
\begin{codex}
? - loop 0 X?
ANSWER: {'X': 0}
ANSWER: {'X': 1}
...
\end{codex}

\section{Natlog as an  Orchestrator for Neuro-Symbolic Deep Learning Systems}\label{jax}

We will show next how the use of Natlog as an embedded logic-based scripting language can simplify
the design and the execution of neural networks as well as making their
internal logic easily understandable.

\subsection{Accessing the namespaces of the Python packages}
To ensure unimpeded access to relevant Python objects that can be as many as a few hundred for packages like JAX or Pytorch, we have devised a namespace sharing mechanism with the Python side. For objects visible in the original namespace of the ``{\tt natlog}'' package this can be achieved by calling {\tt eval} on the name of the function or generator. However, when Natlog itself is imported as a package (as in the JAX and Pytorch apps relying on it) we collect to a dictionary the names visible in the Python client importing Natlog in the app and
then pass it to Natlog. 

\subsection{Interoperation with JAX }

A natural partner to logic-based languages when interacting with deep learning systems is a declaratively designed neural-network like Google and DeepMind's JAX \cite{jax}.

JAX is referentially transparent (no destructive assignments) and fully compositional, via a lazy application of matrix/tensor operations, automatic gradient computations and a just-in-time compilation function also represented as a first-class language construct.

The interaction with a logic-based programming language is facilitated by the lazy functional syntax of JAX (seen as an embedded sublanguage in Python).

After defining a set of data loaders, initialization functions and basic neural network layers, both learning and inference can be expressed in JAX as a composition of functions (more exactly as a {\em future} consisting of such a composition to be eventually activated after a compilation step).

In a Python-based logic language like Natlog, this orchestration process can be expressed as a set of Horn Clauses with logic variables bound to immutable JAX  objects transferring inputs and outputs between neural predicates representing the network layers. JAX's high-level, referentially transparent matrix operations can be reflected safely as Natlog predicates with the result of the underlying N-argument function unified with their N+1-th extra argument. As such, they can be passed as bindings of logic variables between clause heads and clause bodies in an easy to understand, goal-driven design and execution model.

JAX's equivalents of Natlog's compound terms, called {\em pytrees}, hold arbitrary aggregates of data and can be differentiated with  JAX's {\tt grad} operator as a single unit.

Hyperparameter optimization searches can be naturally expressed as constraint-driven optimization processes.

The synergy between the declarative neural framework and the declarative logic orchestrator can also help with identifying the complex causal chains needed for debugging, optimizing the network architecture as well as with the explainability of both the design and the execution of the resulting neuro-symbolic system.

The following Natlog code snippet\footnote{see \url{https://github.com/ptarau/natlog/blob/main/apps/deepnat/natjax.nat} for  more details}
generates a ``deep xor'' dataset, known to be unusually challenging even for deep neural nets. 
\begin{code}
xor 0 0 0.
xor 0 1 1.
xor 1 0 1.
xor 1 1 0.
\end{code}
The predicate {\tt iter} recurses $N$ times over the truth table of {\tt xor} to obtain the truth table of size $2^N$ of $X_1 ~xor~ X_2 ~xor~ ... X_n$ that
we will use as our synthetic dataset.

\begin{code}
iter N Op X Y: iter_op N Op () E 0 Y, to_tuple E X.
\end{code}
\begin{code}
iter_op 0 _Op E E R R.
iter_op I Op  E1 E2 R1 R3 :
   when I > 0, with I - 1 as J,
   Op X R1 R2,
   with X + X as XX,   
   with XX - 1 as X1,  
   iter_op J Op (X1 E1) E2 R2 R3.
\end{code}

The dataset will be passed to JAX after conversion to tuples of tuples, that will eventually become JAX tensors.
\begin{code}
dataset N Op Xss (Ys): 
  findall (X Y)  (iter N Op X Y) XssYsList,
  to_pairs XssYsList XssList YsList,
  to_tuple XssList Xss,
  to_tuple YsList Ys.

to_pairs () () (). 
to_pairs ((Xs Y)  Zss) (Xs Xss) (Y Ys) : 
   to_pairs Zss Xss Ys.
\end{code}

We also generate, depending on the number of variables {\tt N}, an appropriate list of hidden-layer sizes (via the predicate {\tt hidden\_sizes}) for a Multi-Layer Perceptron that we design  in JAX to be trained on our synthetic dataset. We use the ``\verb~'~'' marker to indicate Python calls to functions like {\tt train\_model} and {\tt test\_model} implemented in a companion Python file\footnote{\url{https://github.com/ptarau/natlog/blob/main/apps/deepnat/natjax.py}}.

\begin{code}
run N Op Seed Epochs Loss Acc LossT AccT:
   dataset N Op Xss Ys,
   split_dataset Xss Ys X Xt Y Yt,
   hidden_sizes N Sizes,
   #print hidden sizes are Sizes,
   `train_model X Y Sizes Epochs (Model LossFun), 
   `test_model Model LossFun X Y (Loss Acc),      
   `test_model Model LossFun Xt Yt (LossT AccT).  
\end{code}
Similar calls to pass the dataset to Python and to split it into ``train'' and ``test'' subsets, as well as seamless interaction with objects like JAX-arrays (seen by Natlog as constant symbols) are shown in the Natlog definition of {\tt split\_dataset}.

\begin{code}
split_dataset Xss Ys Xtr Xt Ytr Yt:
   `array Xss X,
   `array Ys Y0,
   `transpose Y0 Y,
   `split X Y 0 0.1 (Xtr Xt Ytr Yt),
   show_sizes (X Y Xtr Xt Ytr Yt).
\end{code}
For instance, ``\verb~`array Xss X~'' converts a Natlog tuple of tuples {\tt Xss} to a two-dimensional JAX array {\tt X}, relying on the fact that JAX's underlying {\tt numpy} matrix library does such operations automatically.
Note that the embedding of Natlog in Python makes such data exchanges $O(1)$ in time and space as no data-conversions need to be performed. Note also that the same applies to compound terms that correspond one-to-one to immutable nested tuples in Python and in particular to {\tt pytrees}, instrumental in creating more advanced neural nets in the JAX ecosystem. 

\subsection{Interoperation with Pytorch}
In the {\tt Pytorch} ecosystem a 
combination of object-orientation and callable models encapsulate the underlying complexities of  dataset management,  neural network architecture choices, automatic differentiation, backpropagation and optimization steps.
However, we can follow a similar encapsulation of architectural components as we have shown for JAX, and delegate to Python the details of building and initializing the network layers\footnote{see \url{https://github.com/ptarau/natlog/blob/main/apps/deepnat/nattorch.py}} with Natlog used to glue together the training and inference steps\footnote{see \url{https://github.com/ptarau/natlog/blob/main/apps/deepnat/nattorch.nat}}.

\section{Logic Grammars as Prompt Generators}\label{prompts}

We will next overview  Natlog applications for text-to-text and text-to-image generation. 
We refer to the Natlog code\footnote{see \url{https://github.com/ptarau/natlog/blob/main/apps/natgpt/chat.nat}} and its Python companion\footnote{see \url{https://github.com/ptarau/natlog/blob/main/apps/natgpt/chat.py}} for  full implementation details.

\subsection{Prompt engineering by extending GPT3's text completion}

GPT3 is basically a text completion engine, which, when given  an initial segment of a sentence or paragraph
as a {\em prompt}, it will complete it, often with highly coherent and informative results.

Thus,  to get from GPT3 the intended output (e.g., answer to a question,  elations extracted from a sentence, building analogies, etc.) one needs to rewrite the original input into a prompt that fits GPT3's text completion model.

We will use here Natlog's syntactically lighter Definite Clause Grammars, with one or more terminal 
symbols prefixed by ``\verb~@~'' and ``\verb~=>~'' replacing Prolog's ``\verb~-->~''.
A prompt generator with ability to be specialized for several ``kinds'' of prompts is described by the DCG rule:
\begin{code}
prompt Kind QuestText => prefix Kind, sent QuestText, suffix Kind.
\end{code}
The predicate {\tt sent} takes a question sentence originating from a user's input and maps it
into a DCG non-terminal transforming cons-list {\tt Ws1} into cons-list {\tt Ws2}:
\begin{code}
sent QuestText Ws1 Ws2 : 
   `split QuestText List, to_cons_list List Ws, append Ws Ws2 Ws1.
\end{code}
The predicate {\tt query} takes the DCG-generated {\tt prompt} derived from user question {\tt Q}
and converts it back to a string passed to GPT'3 completion API by a call to the function {\tt complete}, implemented in Python.
\begin{code}
query Kind Q A: prompt Kind Q Ps (), to_list Ps List, `join List P, `complete P A.
\end{code}
Next we will describe  specializations to question/answering, relation extraction and analogy invention.

An easy way to transform a question answering task into a completion task is to emulate a 
hypothetical conversation:
\begin{code}
prefix question =>   @ 'If' you would ask me.
suffix question =>  @ 'I' would say that.
\end{code}
Similarly, extraction of subject-verb-object phrases can be mapped to completion tasks as in:
\begin{code}
prefix relation =>
   @ 'If' you would ask me what are the subject and the verb and the object in .
suffix relation =>
   @  'I' would say subject is.
\end{code}
For analogy invention we create a custom trigger as follows: 
\begin{code}
trigger X Y Z => 
   @ given that X relates to Y by analogy 'I' would briefly say that Z relates to.

analogy X Y Z A:
   trigger X Y Z Ps (), to_list Ps List, `join List P, `complete P A.
\end{code}

We will next show  interaction examples for all these use cases. First, question answering:
\begin{codex}
?- query question 'how are transformers used in GPT' R.
ANSWER: {'R': 'transformers are used in GPT (Generative Pre-trained Transformer) 
models  to generate text from a given prompt. The transformer architecture is
used to learn the context of the input text and generate a response based on the
context. GPT models are  used in many natural language processing tasks such as 
question answering, machine translation, summarization, and text generation.'}
\end{codex}

Next, relation extraction. Note that after some preprocessing, the extracted triplets
can be used as building blocks for knowledge graphs.

\begin{codex}
?- query relation 'the quick brown fox jumps over the lazy dog' R.
ANSWER: {'R': '"quick brown fox", verb is "jumps" and object is "lazy dog".'}

?- query relation 'high interest rates try to desperately contain inflation' R.
ANSWER: {'R': '"high interest rates", verb is "try to desperately contain", 
and object is "inflation".'}
\end{codex}
Finally, some examples of analogical reasoning that show GPT3 finding the missing
component and explaining its reasoning. 
\begin{codex}
?- analogy car wheel bird A?
ANSWER: {'A': 'wing by analogy. This is because both car and wheel are used for 
transportation, while bird and wing are used for flight.'}

?- analogy car driver airplane A?
ANSWER: {'A': 'pilot by analogy. The pilot is responsible for the safe operation 
of the airplane, just as the driver is responsible for the safe operation 
of the car.'}

?-  analogy cowboy horse advertiser A?
ANSWER: {'A': 'customer by analogy in that they both need each other to achieve 
a goal. The advertiser needs the customer to purchase their product or service 
and the customer needs the advertiser to provide them with the product or 
service they are looking for.'}
\end{codex}

\subsection{Text-to-image with DALL.E}

With magic wands on a lease from generators like  DALL.E \cite{dalle} or Stable Diffusion \cite{stabdif},
Natlog's Definite Clause Grammars can work as easy to customize  prompt generators for such systems.

As the same OpenAI API (with a slightly different Python call) can be used for
text-to-image generation (followed by displaying the generate picture in the user's default browser),
the interaction with Python is expressed succinctly by the predicate {\tt paint} that receives
as {\tt Prompt} the description of the intended picture from the user.
\begin{code}
paint Prompt: `paint Prompt URL, #print URL, #browse URL.
\end{code}

The query to visualize in the user's browser one of the DCG-generated prompts is:
\begin{codex}
?- paint '<text description of intended image>'.
\end{codex}
with some detail delegated to Python and taking advantage of the fact that the same OpenAI API is used for both text-to-text and text-to-image generation.

The Natlog DCG, in generation mode, can iterate over possible styles and content elements of a desired painting as in the following example:
\begin{code}
image => style, subject, verb, object.

style => @photorealistic rendering.
style => @a dreamy 'Marc' 'Chagall' style picture.
style => @an action video game graphics style image.

subject => @of, adjective, noun.
noun => @robot.
verb => @walking.
adjective => @shiny.

object => location, @with,  instrument.

location => @on planet 'Mars'.
instrument => @high hills and a blue purse.
instrument => @a sombrero hat.

\end{code}

This generates text ready to be passed via the OpenAI Python APIs to DALL.E:

\begin{codex}
?- image Words (), `to_tuple Words Ws, #writeln Ws, nl, fail.
photorealistic rendering of shiny robot walking on planet Mars with high hills 
and a blue purse 
photorealistic rendering of shiny robot walking on planet Mars with a sombrero hat
..... 
\end{codex}

Besides the expected dependence on the {\tt style} component (photorealistic vs. Chagall-style), 
as an illustration of GPT3's stereotyping bias, female and respectively male features would  be
derived from the generated robot pictures depending on the {\tt purse} vs. {\tt sombrero hat}
picked by the DCG, 
as can be seen in the generated images\footnote{at 
\url{https://github.com/ptarau/natlog/tree/main/apps/natgpt/pics}}.

\section{Related Work}\label{rel}

An introduction to Natlog, its initial proof-of-concept implementation and its content-driven indexing mechanism are covered in \cite{iclp21}, but the language constructs and application discussed in this paper are all part of a fresh, ``from scratch'' implementation.
We have implemented similar First-Class Logic Engines in the BinProlog  \cite{tarau:cl2000} and Jinni Prolog systems \cite{tarau:shaker}, but their addition to Natlog, is  motivated by their strong similarity with Python's own coroutining mechanisms. The use of coroutining in languages like {\tt C}\#, JavaScript and Python has also been used in the Yield Prolog system \cite{yieldprolog} as a  facilitator for implementing backtracking, similarly to our implementation. Contrary to Natlog, which adopts its own surface syntax and reflection-based interaction with the host language, Yield Prolog requires idiomatic use of the syntax of the target language, making it significantly more cumbersome to work with. 

Interoperation with Python has been also used in Problog \cite{de2007problog} and DeepProblog \cite{deepproblog}, in the latter as a facilitator for neuro-symbolic computations.
A comprehensive overview of neuro-symbolic reasoning, including logic-based, term-rewriting and graph-based  is given in \cite{neurosym}. While in \cite{iclp21} we describe, as an example of neuro-symbolic interaction the use of a neural network as an alternative multi-argument indexer for Natlog, in this paper our focus is on the use of Natlog as an orchestrator for putting together and training deep learning systems and as a prompt generator for Large Language Models.

OpenAI's own {\tt GPT 3.5}-based {\tt ChatGPT}\footnote{https://chat.openai.com/chat} automates the mapping of more queries (e.g., questions, code generation, dialog sessions, etc.) using an extensive 
Reinforcement Learning With Human Advice process \cite{gptrl}. 
By contrast, our DCG-supported approach relies exclusively on the pure GPT3 text-completion API on top of which we engineer task-specific prompts.

\section{Conclusion}\label{conc}

Motivated by expressiveness challenges faced by logic-based programming languages in the context of today's competitive landscape of alternative paradigms as well as from  neural net-based machine learning frameworks, we have  sketched  some implementationally speaking ``low-hanging'' enhancements to them, with  emphasis on coroutining methods and neuro-symbolic interoperation mechanisms. The main contributions of this work, likely to be reusable when bridging other Prolog systems to deep-learning ecosystems, are techniques that facilitate interoperation in the presence of high-level language constructs like finite functions, generators, on-demand computations, backtracking, nested immutable data types and strong reflection and metaprogramming features. The use cases described in the paper show the practical expressiveness of the Natlog-Python symbiosis by enhancing interaction with today's latest generation deep-learning tools with the declarative convenience of a lightweight embedded logic programming language.

\bibliographystyle{eptcs}

\bibliography{tarau,ml,proglang}

\end{document}

%% file: pheader22.tex
\usepackage{graphicx}
\usepackage{mathptmx}
\usepackage{tikz-qtree}
\usepackage{filecontents}
\usepackage{csvsimple}
\usepackage{amsfonts,amsmath}
\usepackage{subfig} 
\usepackage{url}
\usepackage{verbatim}
\usepackage{color}
\usepackage{amsmath} 
\usepackage{proof}

\renewcommand{\phi}{\varphi}

\input prolog.tex

\definecolor{lgray}{gray}{0.95}
\definecolor{lblue}{rgb}{0.90,0.90,1.00}
\definecolor{lyellow}{rgb}{1.00,1.00,0.70}

\usepackage{listings}
\newtheorem{ex}{Example}

\lstnewenvironment{codex}
    {\lstset{}%
      \csname lst@SetFirstLabel\endcsname}
    {\csname lst@SaveFirstLabel\endcsname}
\newcommand{\BI}[0]{\begin{itemize}}
\newcommand{\EI}[0]{\end{itemize}}
\newcommand{\I}[0]{\item}
\newcommand{\BE}[0]{\begin{enumerate}}
\newcommand{\EE}[0]{\end{enumerate}}

\newcommand{\BX}[0]{\begin{ex}}
\newcommand{\EX}[0]{\end{ex}}
\newcommand{\BV}[0]{\begin{verbatim}}
\newcommand{\EV}[0]{\end{verbatim}}

\newcommand{\BF}[0]{\begin{filecontents*}{data.csv}}

\newcommand{\BQ}[0]{\color{blue}\begin{quote}}
\newcommand{\EQ}[0]{\end{quote}\color{black}}

\def \bscale1 {0.25}
\def \bscale {0.25}

\newcommand{\FIG}[4]{
\begin{figure}[htbp]
\centering
{\includegraphics[scale=#3]{#4}}
\caption{#2}
\label{#1}
\end{figure}
}

